\documentclass[conference]{IEEEtran}
\IEEEoverridecommandlockouts
\usepackage{cite}
\usepackage{makecell}
\usepackage{amsmath,amssymb,amsfonts}
\usepackage{algorithmic}
\usepackage{graphicx}
\usepackage{textcomp}

\usepackage{fancyhdr}

\fancypagestyle{firstpage}{
    \fancyhf{}
    \fancyhead[L]{Published in Proceedings of the 2024 IEEE Cyber Science and Technology Congress (CyberSciTech)} 
}

\usepackage{balance}

\usepackage{xcolor}
\usepackage[numbers]{natbib}
\usepackage{subcaption}
\def\BibTeX{{\rm B\kern-.05em{\sc i\kern-.025em b}\kern-.08em
    T\kern-.1667em\lower.7ex\hbox{E}\kern-.125emX}}

\begin{document}

\title{NCAA Bracket Prediction Using Machine Learning and Combinatorial Fusion Analysis
}

\author{\IEEEauthorblockN{ Yuanhong Wu}
\IEEEauthorblockA{\textit{\makecell{Department of Computer and  Information \\ Science }}\\
\textit{Fordham University}\\
New York, NY, 10023, USA \\
ywu463@fordham.edu} 

\and
\IEEEauthorblockN{ Isaiah Smith}
\IEEEauthorblockA{\textit{Gabelli School of Busines} \\
\textit{Fordham University}\\
New York, NY, 10023, USA \\
is5@fordham.edu}
\and
\IEEEauthorblockN{Tushar Marwah}
\IEEEauthorblockA{\textit{Gabelli School of Busines} \\
\textit{Fordham University}\\
New York, NY, 10023, USA \\
tm2@fordham.edu} 

\and
\IEEEauthorblockN{Michael Schroeter}
\IEEEauthorblockA{\textit{Gabelli School of Busines} \\
\textit{Fordham University}\\
New York, NY, 10023, USA  \\
mschroeter2@fordham.edu}
\and
\IEEEauthorblockN{Mohamed Rahouti}
\IEEEauthorblockA{\textit{\makecell{Department of Computer and Information  \\ Science}} \\
\textit{Fordham University}\\
New York, NY, 10023, USA  \\
mrahouti@fordham.edu}
\and
\IEEEauthorblockN{D. Frank Hsu}
\IEEEauthorblockA{\textit{\makecell{Department of Computer and Information  \\ Science}} \\
\textit{Fordham Universiyt}\\
New York, NY, 10023, USA\\
hsu@fordham.edu}
}

\maketitle
\thispagestyle{firstpage}
\begin{abstract}
Machine learning models have demonstrated remarkable success in sports prediction in the past years, often treating sports prediction as a classification task within the field. This paper introduces new perspectives for analyzing sports data to predict outcomes more accurately. We leverage rankings to generate team rankings for the 2024 dataset using Combinatorial Fusion Analysis (CFA), a new paradigm for combining multiple scoring systems through the rank-score characteristic (RSC) function and cognitive diversity (CD). 
Our result based on rank combination with respect to team ranking has an accuracy rate of $74.60\%$, which is higher than the best of the ten popular public ranking systems ($73.02\%$). This exhibits the efficacy of CFA in enhancing the precision of sports prediction through different lens.
\end{abstract}


\section{Introduction}
The NCAA Men's Basketball Championship, also known as March Madness, is an annual event that brings together the best college basketball teams in the nation. Each year, 68 teams from the NCAA's Division I compete in this high-stakes, single-elimination tournament. The excitement builds over three weeks, with each game drawing millions of viewers and culminating in the crowning of the national champion. The unpredictability of each game, with potential for upsets and Cinderella stories, adds to the allure and drama of the competition, making March Madness a beloved tradition in American sports \cite{logan2018cinderella}. 

Predicting the outcome of the NCAA men's basketball tournament has gained much attention in recent years. In the decades since, filling out a March Madness bracket has become an American sports staple. In May 2023, the NCAA released the results from a survey of 3,527 eighteen-to-twenty-two year olds. It found that sports wagering is pervasive, with $58\%$ having engaged in at least one sports betting activity \cite{rychlak2023mobile}.  

In recent years, the intersection of sports and computational science has emerged as a compelling arena for
research and innovation. The utilization of machine learning methods to predict sports outcomes has
garnered significant attention due to its potential to revolutionize how we perceive and engage with
athletic events. Even though machine learning models can provide valuable insights into sports games, we note that sports prediction has an inherently unpredictable nature \cite{bunker2022application}.  Despite extensive statistical analysis and sophisticated models, factors such as player injuries, team chemistry, coaching strategies, and even external variables like weather conditions can significantly impact the result of a game. Moreover, the emotional and psychological aspects of sports, such as momentum swings and underdog performances, further complicate prediction tasks.

This paper aims to apply several ML-based models and CFA in the bracket prediction of the NCAA men's basketball tournament. Specifically, we select five models as the base models to generate a large pool of ensemble models using Combinatorial Fusion Analysis (CFA). CFA is a paradigm that seeks to provide methods and workflows for combining multiple scoring systems in computational learning and modeling, informatics, and intelligent systems \cite{hsu2024combinatorial, hsu2006combinatorial}.
 We utilize the ensemble performance across the previous 10 years to show improvement in the \textquotedblleft future" data for 2024. Both game ranking and team ranking are obtained through the optimal ensemble model.

The structure of this paper is as follows. We include some relevant literature on NCAA prediction in Section II. Section III focuses on the CFA framework. Section IV is devoted to feature selection and introduction of the five base models. Results of the prediction are included in Section V. Finally in section VI,  discussion and conclusion are given.

\section{Previous work}

Throughout the years, academic researchers and basketball statisticians have attempted to predict the outcomes of March Madness games using a wide variety of explanatory variables and empirical methods. Some of the earliest analyses used the seed assigned to each team as a rather intuitive means of predicting the likelihood that a team would win and the margin of doing so \cite{boulier1999sports, caudill2003predicting, smith1999can}. The collective findings of these initial studies indicated that an obvious advantage exists for a team that possesses a higher seed than its regional opponents. However, subsequent analyses showed that the relationship between seed and performance breaks down as the tournament progresses \cite{jacobson2009seeding}.

To this end, additional studies have investigated a variety of predictors other than \textquotedblleft seeding". These variables have ranged from regular season winning percentages, margins of victory in regular season games, records against tournament teams, and Vegas point spreads, to ratings (e.g., NCAA RPI measures; KenPom, Sagarin, and Massey ratings) that are specific to an organization or website  \cite{hoegh2015nearest}. Several studies employ methods that simultaneously incorporate many of these variables in their models. Others compare the predictability of these measures and models in order to determine which ones are most accurate. The logit regression/Markov-chain model developed by \cite{kvam2006logistic}, for example, incorporated variables such as margin of victory, game location, and strength of schedule into its metrics and was found to be more predictive than Vegas betting odds and other common ranking systems.  Brown and Sokol \cite{brown2010improved}
presented an improved LRMC model where they replaced
logistic regression with two separate empirical Bayes
models: one is to predicate the outcome of individual
games on neutral courts like the NCAA Tournament, and
the other is to directly identify the best teams. They
found that the improved LRMC model with logistic
regression replaced by either one of Bayes models had
better performance. 

Yuan et al. \cite{yuan2015mixture} combined several previous models
-- logistic regression with backward elimination, L2
regularization, L1 regularization respectively and
stochastic gradient boosting, neural networks, and
ensemble models which combined several models
 mentioned above. Among all these models, logistic
regression, stochastic gradient boosting, and neural
network algorithms obtained the best performance.
However, the combined ensemble models did not outcompete
the individual models.

\section{Combinatorial Fusion Analysis}

Ensemble models have received much attention in sports prediction due to their ability to improve predictive accuracy and robustness by combining multiple models. These advanced techniques effectively capture complex patterns and reduce the risk of overfitting, making them highly suitable for the dynamic and multifaceted nature of sports data. Combinatorial Fusion Analysis (CFA), proposed by Hsu, Shapiro, and Taksa  \cite{hsu2002methods} and Hsu, Chung and Kristal \cite{hsu2006combinatorial} in the early 2000s, is a new ensemble paradigm that provides methods and workflow for combining multiple scoring systems using rank-score characteristic (RSC) function and cognitive diversity (CD) \cite{hsu2006combinatorial, hsu2024combinatorial}.   CFA  not only can work in Euclidean space like most other ensemble models do but is able to leverage ranking information of the data for combining base models. CFA has been widely employed in various disciplines, including bioinformatics \cite{brown2012chip}, virtual screening and drug discovery \cite{jiang2023enhancing}, portfolio management \cite{wang2019improving}, and information retrieval \cite{frank2005comparing}, among others.

\subsection{Multiple scoring systems}
 A scoring system consists of a score function, a derived rank function, and the corresponding rank-score characteristic (RSC) function. Let $D = \{d_1, d_2, \dots, d_n  \}$ be a set of $n$ data items. A score function $s_A$ is a function that assigns a single numerical number in $\mathcal{R}$ to each data item in $D$. For classification tasks, the score function can be the probabilities from the output of a classifier. A rank function $r_A$ is obtained by sorting the score values in the score function into descending order and assigning the rank order of that score value to the data item. Hence $r_A(d_i) \in \mathcal{N}$, where $N=\{1, 2, \dots, n \}$ and $n$ is the cardinality of $D$. After determining the score function and the rank function, an RSC function $f_A$ can subsequently be derived as, $\text{ for } i \in \{1, \dots, n \}$,
 \begin{equation}
      f_A(i) = \big(s_A \circ r_A^{-1}\big)(i) = s_A\big(r_A^{-1}(i)\big), 
 \end{equation}

Figure 1(a) presents the relationship among the score function, rank function, and RSC function \cite{hsu2006combinatorial,
hsu2024combinatorial}. Note that the RSC function $f_A$ is constructed from natural number $\mathcal{N}$ to real number $\mathcal{R}$, meaning that the RSC function is independent of the data items $D$. This property builds the RSC function's independence across domains in machine learning tasks.  The examples of two RSC functions are drawn in  Figure 1(b). The x-axis is ranks while the y-axis is the corresponding scores for each rank. Clearly, the RSC function is a non-increasing function. Particularly, in the case that a scoring system contains tied score values, the line will exhibit horizontal segments. Hurley et al. \cite{hurley2020multi} indicated that the RSC function of a scoring system characterizes the scoring system analogous to the role played by the cumulative distribution function (CDF) in statistics.

        
        

\begin{figure}[ht]
    \centering
    \begin{subfigure}[b]{0.45\textwidth}
        \centering
        \caption{\phantom{dajfkjasdkafdkfjaksdlfasdfsadfasddsfjkjaskdfljdskfjalskjdfklsjdafkljasdf}}
        \includegraphics[width=\textwidth]{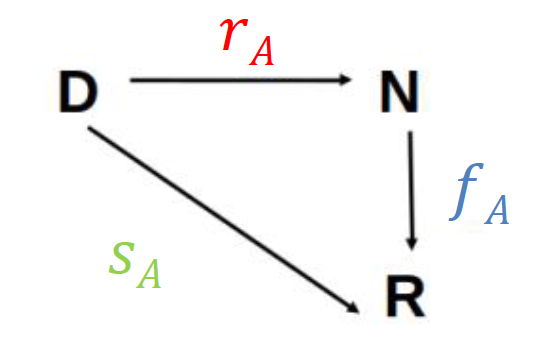} 
    \end{subfigure}
    \hfill
    \begin{subfigure}[b]{0.45\textwidth}
        \centering
        \caption{\phantom{dajfkjasdkafdkfjaksdlfasdfsadfasddsfjkjaskdfljdskfjalskjdfklsjdafkljasdf}}
        \includegraphics[width=\textwidth]{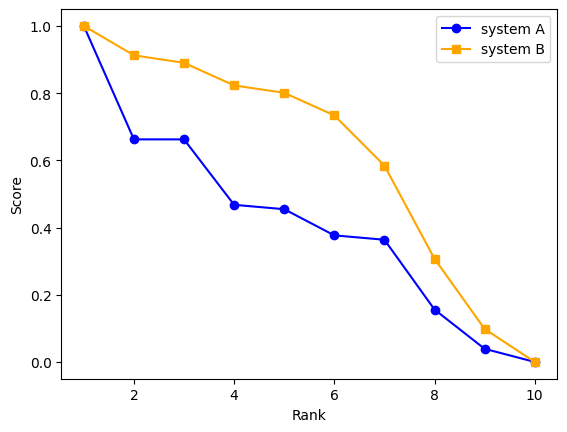} 

    \end{subfigure}
    \label{fig:both_figures}
     \caption{(a) Relationship between score, rank, and RSC function, where $D$ is a set of data items, N and R refer to natural numbers and real numbers, respectively \cite{hsu2006combinatorial,
hsu2024combinatorial}. (b) RSC functions for two scoring systems A and B.}
\end{figure}

\subsection{Cognitive diversity}
Ensemble models have shown superior performance by leveraging the strengths of individual base algorithms and have been broadly employed in various domains and contexts. Many researchers have recognized that the diversity of a set of base classifiers plays a critical role in constructing a successful ensemble model \cite{bian2021does, mao2015weighted, windeatt2005diversity}. However, there is no generally acknowledged definition of diversity. Various diversity measures have been proposed in the literature to enhance ensemble classifier performance, such as Q statistic, Double Fault measure, etc \cite{hurley2020multi, bian2021does, mao2015weighted, windeatt2005diversity}.

Cognitive diversity, as defined by Hsu, Kristal and Schweikert  \cite{hsu2010rank}, is a new type of diversity measure that incorporates rankings of the data items into considerations. It operates independently of the data, as it utilizes the RSC function of the scoring systems. Cognitive diversity
measures the difference between two scoring systems A and B using RSC functions $f_A$ and $f_B$ \cite{hsu2010rank, schweikert2022modeling}. More specifically, cognitive diversity between A and B, $CD(A, B)$, is the distance between RSC functions $f_A$ and $f_B$ \cite{hsu2010rank, yang2005consensus}:

\begin{equation}
    CD(A, B) = \sqrt{\frac{1}{n}\sum_{i=1}^n \bigg(f_A(i) - f_B(i)\bigg)^2}.\label{cd}
\end{equation}
where $n$ represents the number of data items. It's a pair-wise diversity measure that captures dissimilarity between two scoring systems. The diversity strength (DS) of algorithm A is the average cognitive diversity between A and all other score systems. Let the ensemble classifier contain $m$ scoring systems, that is, $E = \{C_1, C_2, \dots, C_m\}$, the diversity strength of a scoring system $C_i$ in the ensemble is defined  \cite{jiang2023enhancing, hsu2010rank} as:
\begin{equation}
    DS(C_i) = \frac{\sum_{j\neq i} CD(C_i, C_j)}{m-1}.\label{ds}
\end{equation}
where $m$ denotes the number of scoring systems.

\subsection{Model combination}
One of the two components of building an ensemble model is the way to combine the base classifiers \cite{liu2017combination,
hsu2024combinatorial}. In ensemble learning, popular methods for integrating models are average combination for regression and majority voting for classification. Average combination mitigates individual model errors by computing the mean of their predictions, while majority voting determines the final classification by selecting the class with the highest number of votes from the ensemble members. CFA framework employs three methods to aggregate predictions of base learners: average combination (AC), weighted combination by diversity strength (WCDS), and weighted combination by performance (WCP)\cite{hsu2024combinatorial, hsu2006combinatorial, jiang2023enhancing}. The performance in WCP depends on the evaluation method used, such as accuracy, precision, recall, and area under the ROC curve.

 Each scoring system A comprises a score function $s_A$ and a rank function $r_A$, corresponding to the original scores and derived rankings, respectively. Both score combination and rank combination can be utilized to combine base models through any of the three combination approaches, resulting in six distinct combinations. Let $M$ represent a set of $t$ scoring systems, defined as $ M = \{A_1, A_2, \ldots, A_t\} $, on the dataset \( D = \{d_1, d_2, \ldots, d_n\} \). Consider $h$ as the number of classifiers to combine where $h>1$. And $M_h$ denotes any subset of scoring systems $M$ with the condition $|M_h| = h \leq t$. Since there are $t$ scoring systems, the framework generates $\binom{t}{h}$ different ensembles built by AC method, each of which can be computed as in equation (\ref{sc}).

\begin{equation}
    s_{sc}(d_i) = \frac{1}{h}\underset{A_j \in M_h}{\sum} s_{A_j}(d_i). \label{sc}
\end{equation}
where $s_{sc}(d_i)$ and $s_{A_j}(d_i)$ represent the score function of the score combination and the score function for system $A_j$ on data item $d_i$, respectively \cite{hsu2024combinatorial, hsu2006combinatorial, jiang2023enhancing, hurley2020multi}. 

Similarly, the ensembles using WCDS and WCP is calculated respectively:
\begin{equation}
    s_{sc}(d_i) = \frac{\underset{A_j \in M_h}{\sum}DS(A_j)s_{A_j}(d_i)}{\underset{A_j \in M_h}{\sum} DS(A_j)}\label{ab}
\end{equation}
\begin{equation}
    s_{sc}(d_i) = \frac{\underset{A_j \in M_h}{\sum}P(A_j)s_{A_j}(d_i)}{\underset{A_j \in M_h}{\sum} P(A_j)}. \label{e1}
\end{equation}
where  $DS(A_j)$ and $P(A_j)$  refer to the diversity strength and performance of scoring system $A_j$, respectively.

We substitute score function in equation (\ref{sc}), (\ref{ab}) and (\ref{e1}) with rank function, resulting in rank combination for AC, WCDS, and WCP \cite{hsu2024combinatorial, hsu2006combinatorial, jiang2023enhancing, hurley2020multi}.
\begin{equation}
    s_{rc}(d_i) = \frac{1}{h}\underset{A_j \in M_h}{\sum} r_{A_j}(d_i)
\end{equation}
\begin{equation}
    s_{rc}(d_i) = \frac{\underset{A_j \in M_h}{\sum}\frac{1}{DS(A_j)}r_{A_j}(d_i)}{\underset{A_j \in M_h}{\sum} \frac{1}{DS(A_j)}}
\end{equation}
\begin{equation}
    s_{rc}(d_i) = \frac{\underset{A_j \in M_h}{\sum}\frac{1}{P(A_j)}r_{A_j}(d_i)}{\underset{A_j \in M_h}{\sum} \frac{1}{P(A_j)}}
\end{equation}
where $s_{rc}(d_i)$ is the score function of rank combination for data item $d_i$ while $r_{A_j}(d_i)$ denotes the rank function of scoring system $A_j$.

Figure \ref{fig:dia} presents the CFA combination framework. It consists of three key parts; the left part (A) represents the pre-trained models on the dataset $D$. This paper uses five machine learning models, that is, A: logistic regression, B: Support Vector Machine, C: Random Forest, D: XGBoost, and E: Convolutional Neural Network. The sigmoid activation function is used in the output layer and the ReLu activation function for other layers. The CNN model is optimized using the Adam optimizer with a cross-entropy loss function. A brief introduction about these models is given in section IV (c). The middle part of the framework describes the mechanism of CFA generating models. It consists of $(2^t-1-t)$ combined models each of which has six possible forms of combinations: 2 (SC vs. RC) and 3 (AC, WCP, and WCDS). There are various methods to evaluate model performance, such as accuracy, precision at top $k$ (precision $@ k$), AUROC (area under the ROC curve), and AUPRC (area under the precision-recall curve). We rank all the ensemble models the CFA produced based on a selected metric and choose the most optimal one as the final model.

\section{Data sets and base models}
\subsection{Data collection}

Since 2011, Kaggle.com has hosted the 'March Machine Learning Mania' competition aimed at predicting March Madness brackets. It offers a substantial dataset comprising historical regular season game results, tournament bracket outcomes, and detailed team statistics. Participants are tasked with calculating probabilities for potential matchups using statistical modeling and machine learning techniques.

\begin{figure}
    \centering
    \includegraphics[width=\linewidth]{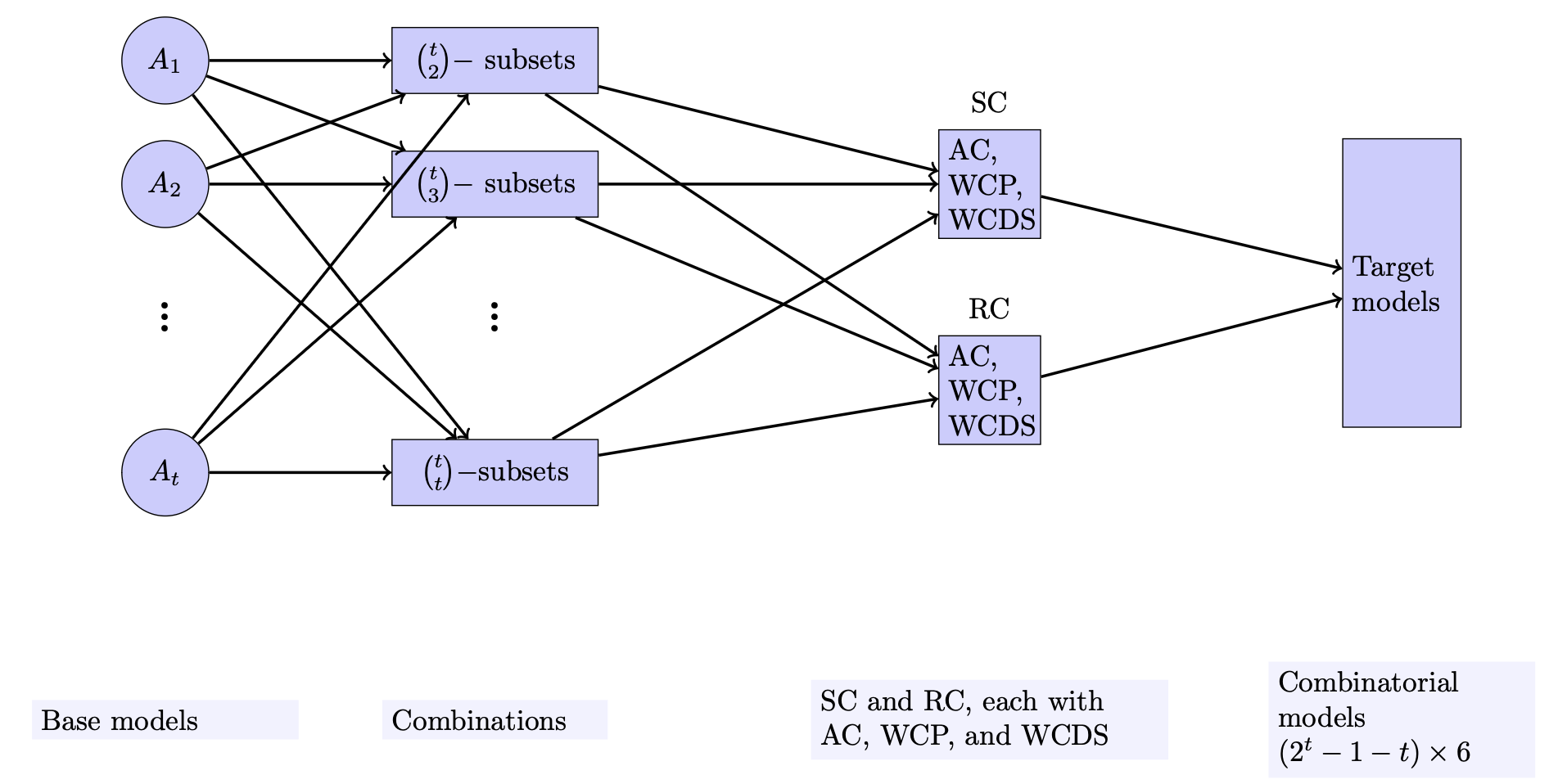}
    \caption{The CFA combination framework for both score and rank combinations, where SC and RC are score combination and rank combination, respectively; AC, WCP, and WCDS refer to average combination, weighted combination by performance, and weighted combination by diversity strength, respectively}
    \label{fig:dia}
\end{figure}
 This annual competition remains active, excluding the year 2020 due to COVID-19. Data from tournaments spanning 2001 to 2022, excluding 2020, are obtained. Additional team data is sourced from the "KenPom" website \footnote{https://kenpom.com}, renowned for its comprehensive statistical analyses for college basketball teams. The website not only provides specific team statistics but also indicates relative team strengths, with lower ranks denoting superior performance in specific aspects. Through feature selection analysis, it is observed that certain variables' ranks are more predictive of game outcomes than their corresponding rating values. Columns with missing data are subsequently removed from the dataset.

 Our dataset includes annual statistics for each team invited to the tournament. Our objective is to predict the winning team in each game. To achieve this, we calculate the difference between the corresponding features of Team 1 and Team 2 for each game. If Team 1 defeats Team 2, we assign a target variable of 1 to the game; otherwise, we assign a target variable of 0. Since Team 1 consistently emerges as the winner in the dataset we have gathered, all games are labeled with a target variable of 1. Consequently, the data only exhibit a single class, rendering it unsuitable for any classification models. To introduce a label 0, we interchange the variables of Team 1 and Team 2 while maintaining their original designations. Therefore, the variables initially attributed to Team 1 now reflect those of Team 2 after swapping. We compute difference variables by subtracting Team 2's variables from Team 1's. Then, we discard all individual features and retain only these difference variables as our initial dataset.


\subsection{Feature selection}

Random Forest, proposed by Leo Breiman \cite{breiman2001random}, is a versatile machine-learning algorithm renowned for its robustness and accuracy in both classification and regression tasks. Random Forest is made up of several decision trees, each decision tree will be full growth without pruning. In addition to classification and regression, Random Forest approach is also widely used for feature selection. 

After we obtain the importance of each feature from the random forest, we perform feature selection using Recursive Feature Elimination with Cross-Validation (RFECV) which recursively removes features, evaluating their performance with cross-validation, to find the optimal subset of features for a given scoring metric. Here we implement RFECV algorithm with 5-fold cross-validation based on log loss metric. We then obtain an optimal subset of 26 features from the original 44 features as the input data set to the machine learning base models.

Team statistics provide a quantitative measure of the team's performance over time. They help coaches, players, and management assess strengths, weaknesses, and areas for improvement. The statistics we selected primarily consist of four parts: offensive efficiency, defensive efficiency, strength of schedule and luck. We then briefly explain what these four types of statistics measure for a team. Offensive efficiency in sports measures a team's effectiveness in converting possessions into points, reflecting their ability to score efficiently. Defensive efficiency, on the other hand, evaluates how well a team prevents opponents from scoring, considering metrics like points allowed per possession and defensive statistics such as rebounds and turnovers. The strength of schedule quantifies the difficulty of opponents faced over a season or tournament, indicating the competitiveness of a team's matchups and the quality of their competition.

In any competition, unforeseen factors such as weather conditions, referee decisions, or unpredictable bounces of the ball can sway the course of a game. These moments of chance can either benefit or disadvantage a team, adding an element of unpredictability and excitement to sports. Many researchers have commented that luck plays an important role for game outcomes In \cite{owen2014measurement, fort2003competitive},  it was indicated that sports outcomes are a mix of skills with good and bad luck.

\subsection{Base models}

Many researchers combined the predictions of multiple classifiers to produce a single classifier which is more accurate than any of the individual classifiers \cite{opitz1999popular}. Two popular ensemble methods are bagging \cite{breiman1996bagging} and boosting \cite{freund1996experiments} which work in Euclidean space. In this paper, we use Combinatorial Fusion Analysis (CFA) proposed by  Hsu, Shapiro, and Taksa and Hsu, Chung and Kristal \cite{hsu2002methods, hsu2006combinatorial}, which incorporates ranking into the models and combines models in both Euclidean and rank space. We have covered the CFA framework in section III. Here we describe the five base models which are the inputs to the CFA framework.

\subsubsection{Model A: Logistic regression}



The logistic regression model assumes a binomial distribution for a binary
response alongside a logit link function. Let $n$ denote the
number of Bernoulli trials that constitute a particular
binomial observation and $N$ denote the number of observations.
Let $y_i , i = 1, \cdots, N$, be the proportion of “successes” out of $n_i$ independent Bernoulli trials, so $n_i y_i \sim Bin(n_i , \pi_i)$,
with $E[y_i] = \pi_i$ independent of $n_i$. Let $x_{ij} , i = 1, \cdots, N, j = 1, \cdots,
p$, be the $j$th predictor for observation $i$. The logistic regression
model is then expressed as \cite{ruspriyanty2018analysis}:
$$
\log\frac{\pi_i}{1-\pi_i} = \sum_{j = 1}^{p} \beta_{j}x_{ij}
$$

To prevent overfitting and enhance model generalization, regularized logistic regression is used, typically employing L1 or L2 regularization. L1 regularization promotes sparsity by shrinking some coefficients to zero, while L2 discourages large coefficients by penalizing their squared values. We use a randomized search technique with 10-fold cross-validation to find the best regularization method and other parameters.

\subsubsection{Model B: Support vector machines}
Support Vector Machines (SVM) aims to find the hyperplane that best separates the data points with different classes in a high-dimensional space. This hyperplane serves as the decision boundary that maximizes the margin between the classes. In support vector classification, the separating function can be expressed as a linear combination of kernels associated with the support vectors as \cite{kecman2005support}:
$$
f(x_i) = \underset{x_j\in S}{\sum} \alpha_j y_jK(x_j, x_i) + b
$$
where $x_i$ denotes data points, $y_i \in \{+1, -1\}$ denotes the corresponding class labels and $S$ is the set of support vectors.

SVM can effectively manage non-linearly separable data through the use of kernel functions.  Commonly used kernel functions include Polynomial, Radial Basis Function (RBF), and Sigmoid.  We use a randomized search method to find the optimal kernel function and other parameters based on the log loss function.


\subsubsection{Model C: Random forest}
Random forest \cite{breiman2001random}, has become a cornerstone in modern machine learning due to its efficiency and robustness in handling large, complex datasets. This ensemble learning method constructs and aggregates multiple randomized decision trees to form a strong predictor, performing well even with small sample sizes and high-dimensional data. Lin et al. \cite{lin2014predicting} implemented 5 different supervised learning classification models to predict the winner of National Basketball Association (NBA) games. Through their findings, Random forest was shown the highest accuracy rate of $65.16\%$ among the other four models.

\subsubsection{Model D: XGBoost}

XGBoost, or Extreme Gradient Boosting, is an advanced machine learning algorithm that builds an ensemble of decision trees in a sequential manner. Each tree is trained to correct the errors made by the previous trees, effectively boosting the model's performance. The core idea of XGBoost is to minimize a loss function $L$
 using gradient descent, where the prediction $\hat{y}$ for a given input 
$x$ is represented as a sum of functions from each tree: $\hat{y} = \sum_i^N f_i(x)$ where $f_i$ represents the
$i$th tree, and $N$ is the total number of trees. To prevent overfitting and ensure generalization, XGBoost includes a regularization term $\Omega(f)$ in its objective function $\mathrm{O}$ \cite{wang2019xgboost}:
$$
\mathcal{O} = \sum_{i=1}^nL(y_i, \hat{y_i}) + \sum_{k = 1}^K \Omega(f_k)
$$
where $\Omega(f)=\gamma T+\frac{1}{2}\lambda \sum_{j=1}T w_j^2$. In this formula, $T$ is the number of leaves in the tree, $w_j$ are the leaf weights, 
$\gamma$ is the complexity cost per leaf, and $\lambda$ is the L2 regularization term on leaf weights. 

\begin{figure}[ht]
    \centering
    \begin{subfigure}[b]{0.45\textwidth}
        \centering
        \caption{\phantom{dajfkjasdkafdkfjaksdlfasdfsadfasddsfjkjaskdfljdskfjalskjdfklsjdafkljasdf}}
        \includegraphics[width=\textwidth]{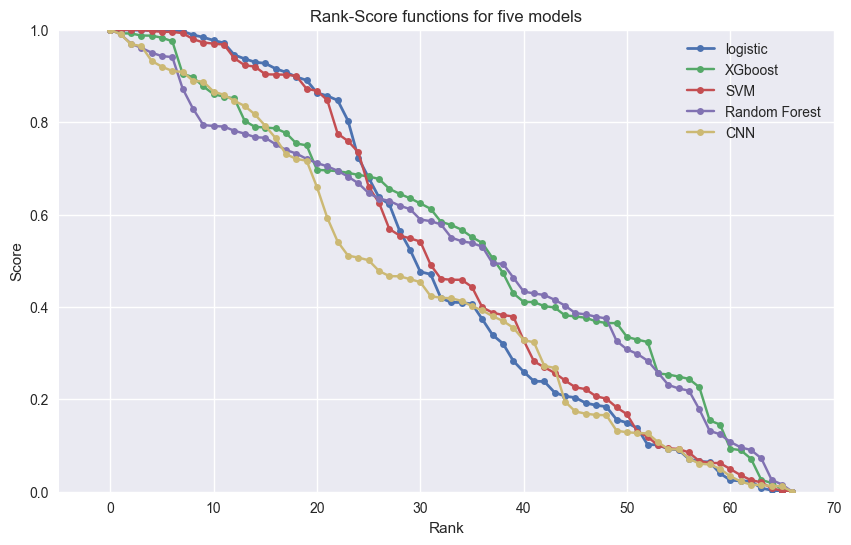} 
    \end{subfigure}
    \hfill
    \begin{subfigure}[b]{0.45\textwidth}
        \centering
        \caption{\phantom{dajfkjasdkafdkfjaksdlfasdfsadfasddsfjkjaskdfljdskfjalskjdfklsjdafkljasdf}}
        \includegraphics[width=\textwidth]{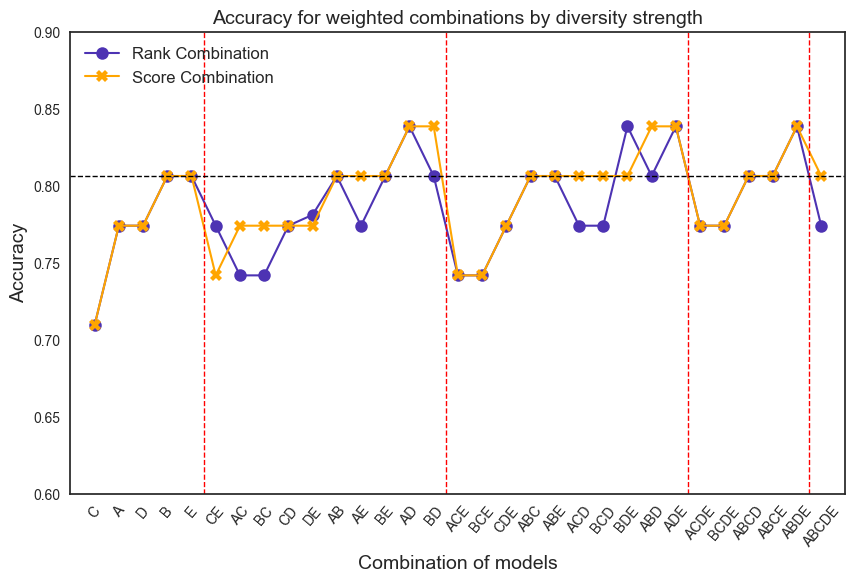} 

    \end{subfigure}
    
    \caption{(a) Five RSC functions  and (b) model combination performance for the year 2022 test data, where A: logistic regression, B: SVM, C: Random
Forest, D: XGBoost, and E: CNN.}
    \label{fig:2022}
\end{figure}

\subsubsection{Model E: Convolutional neural network}
 A Convolutional Neural Network (CNN) 
 is a deep learning model structured to process grid-like data such as images \cite{karn2023machine}. It consists of convolutional layers that apply learnable filters to extract features, pooling layers that reduce spatial dimensions to decrease computational load and prevent overfitting, and fully connected layers that learn complex representations. The final layer typically uses a softmax or sigmoid activation function to produce a probability distribution for classification tasks. See for example Alfatemi et al. \cite{alfatemi2024advancing}.

\section{Results}
\subsection{Model combination and selection}
We employ five base models to construct a large number of ensembles, ensuring robust and reliable predictions. Each base model undergoes training using stratified 10-fold cross-validation with three repetitions, a technique that maintains the class distribution across folds, thereby enhancing the model's ability to generalize across different subsets of the data. To optimize the parameters of each model, we utilize randomized search. This method is computationally efficient compared to an exhaustive grid search, as it samples a fixed number of parameter combinations from a specified distribution. By strategically combining stratified cross-validation and randomized search, we balance the need for thorough model evaluation and parameter tuning while managing computational resources effectively. This approach not only improves the performance and reliability of our models but also ensures that they are well-calibrated and less prone to overfitting.

The five pre-trained models are the input to the CFA framework pipeline in Figure \ref{fig:dia}, leading to $(2^5 - 1 - 5)\times3\times2 = 156$ different ensemble models $t=5$. In order to save computational power and keep it simple, we only investigate diversity strength as the weight for both score combination and rank combination, reducing to $(2^5 - 1 -5)\times2 = 52$ models. The accuracy metric is used as the evaluation method for each generated model. 

Figure \ref{fig:2022} presents the RSC function for five models and the model performance for each possible combination for score combination (orange) and rank combination (blue) on the test dataset of the year 2022. The RSC function shows that the diversity between SVM and logistic regression is smaller than that between SVM and Random Forest. Models exhibiting greater cognitive diversity have a higher chance of a successful ensemble compared to models with lower cognitive diversity, provided their individual performances are relatively good \cite{frank2005comparing}. The model performance for all 52 models is presented in Figure \ref{fig:2022}(b). The black dotted line indicates the best accuracy of the five individual models. It's clear that some combined models show superior performance, such as the \textquotedblleft 2-com" model combining logistic regression (A) and XGboost (D), the \textquotedblleft3-com" model combining logistic regression (A), XGboost (D) and CNN (E) due to good diversity between models.

\begin{figure}[ht]
    \centering
    \begin{subfigure}[b]{0.45\textwidth}
        \centering
        \caption{\phantom{dajfkjasdkafdkfjaksdlfasdfsadfasddsfjkjaskdfljdskfjalskjdfklsjdafkljasdf}}
        \includegraphics[width=\textwidth]{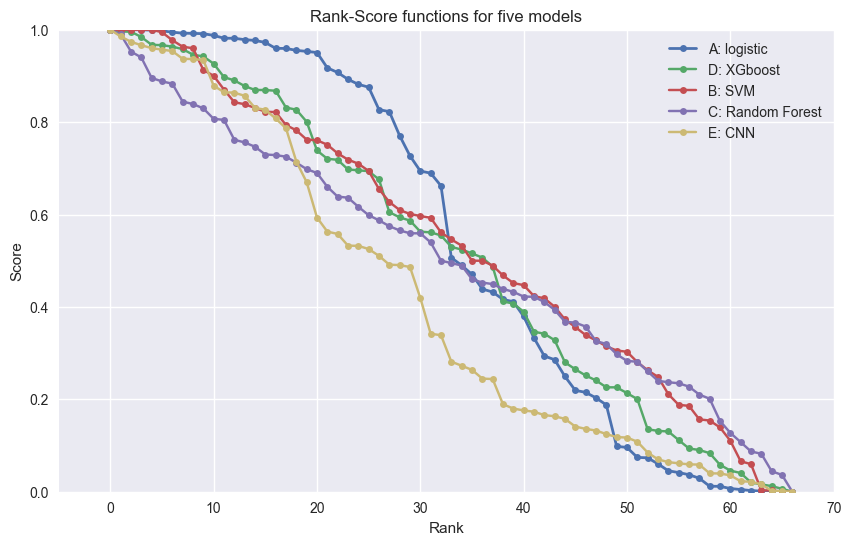} 
    \end{subfigure}
    \hfill
    \begin{subfigure}[b]{0.45\textwidth}
        \centering
        \caption{\phantom{dajfkjasdkafdkfjaksdlfasdfsadfasddsfjkjaskdfljdskfjalskjdfklsjdafkljasdf}}
        \includegraphics[width=\textwidth]{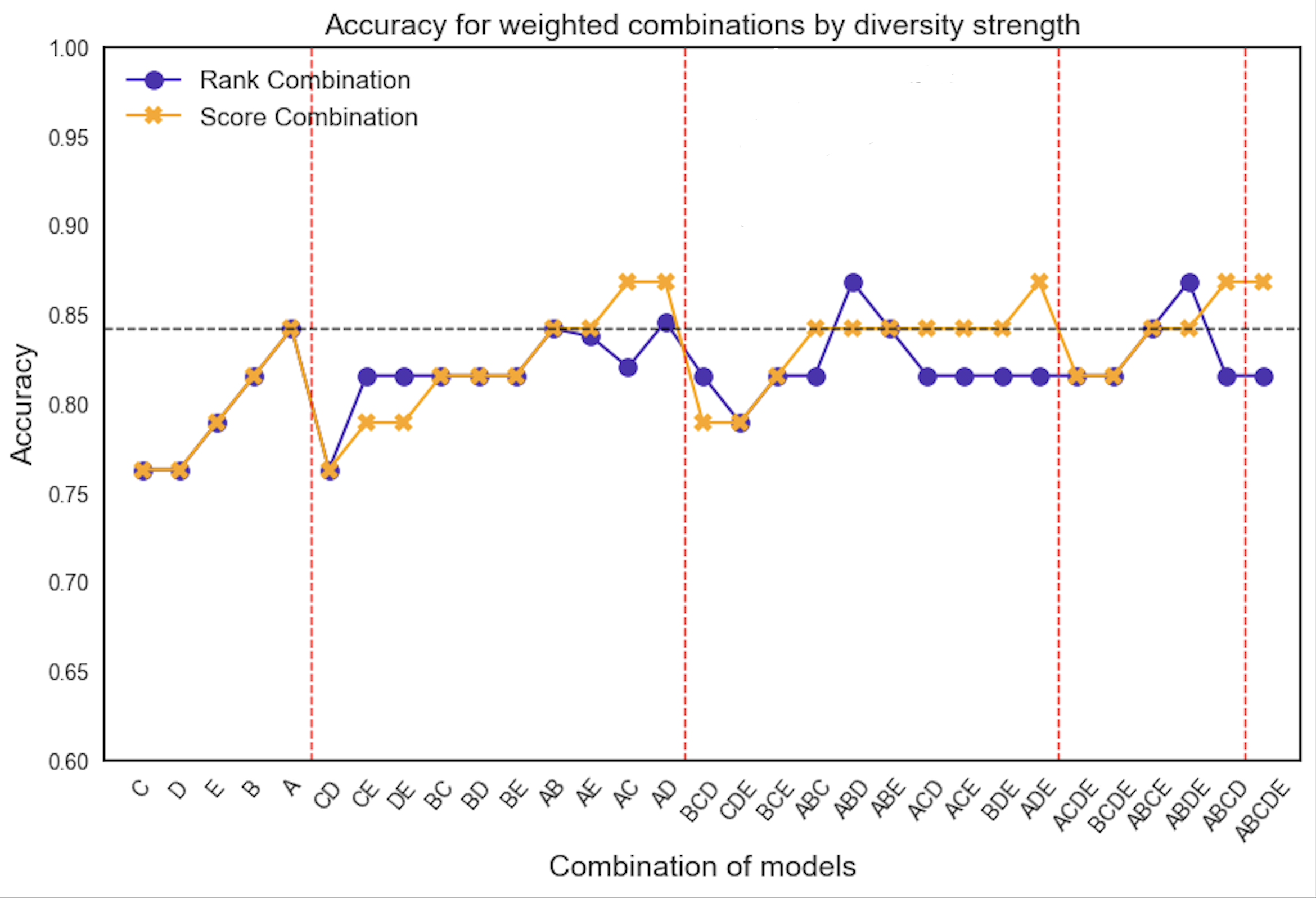} 

    \end{subfigure}
    
    \caption{(a) Five RSC functions and (b) model  combination performance for year 2021 test data, where A: logistic regression, B: SVM, C: Random
Forest, D: XGBoost, and E: CNN.}
    \label{fig:2021}
\end{figure}

Since the CFA framework generates a multitude of models, a question naturally arises: which model should we use on data for the year 2024? We were not able to use the game results in 2024 data to evaluate all the models because in the scenario of this paper, the 2024 NCAA basketball games have not happened yet. The goal is to predict the winning team for each game when the game is played in the field. Therefore, we seek to find patterns from the previous 10 years. In each year, RSC functions for the five base models and accuracy performance for all combinations are drawn for comparison. Figure \ref{fig:2021} presents the results for the 2021 year data. We observe SVM and logistic regression in year 2021 are more diverse than in the year 2022. We record the models that improve the best individual accuracy of the individual model and count the number through the previous 10 years. The combination \textquotedblleft ABE", showing up 6 times, appears to be the most frequent ensemble that shows improvement across the previous 10 years. Hence, the model constructed by logistic regression (A), SVM (B), and CNN (E) is used to perform the subsequent prediction on the future dataset--2024 year data for both rank combination and score combination.

\subsection{Team ranking}
Sport predictions are usually considered as a classification problem by which one class is predicted\cite{prasetio2016predicting}. It was shown \cite{valero2016predicting}  that classification-type models are better than regression-type models when predicting sports game outcomes. In this paper, we view sports prediction as a ranking problem. We provide both perspectives from score combination and rank combination under the CFA framework. We discuss team ranking using the \textquotedblleft ABE" ensemble with rank combination first.

In rank combination, we first obtain a ranking for all the games. Then we convert the game ranking into team ranking to calculate the accuracy. The game ranking considers ranking the confidence of team 1 beating team 2 in each game. A game ranking higher than another game means that team 1 in this game has a higher chance to win its opponent team than team 1 in another game. Subsequently, we calculate the average rank for each of the 64 teams. This average rank serves as the team's score. We then sort these scores in ascending order to establish the team rankings. Finally, we compare these rankings with the actual game outcomes. In our prediction, a team is considered the winner if its rank is ranked higher than that of its opponent. 


Our CFA team ranking using a rank combination framework achieves an accuracy rate of $74.60\%$ which is $1.58\%$ higher than the ten public ranking systems in Table \ref{tab:NCAA} in which NET Rankings and Logan have the highest accuracy of $73.02\%$.

\begin{table}[t]
    \centering
    \begin{tabular}{l|l|l|l}
    \hline
    INCC Stats  & $71.43\%$ & Jelly Juke & $66.67\%$ \\
    Joby Nitty Gritty  & $69.84\%$ & Logan &73.02\% \\
    Massey &  $69.84\%$& Moore & $71.43\% $\\
    NET Rankings & $73.02\%$ & ESPN SOR & $69.84\%$\\
    Sports Ratings & $68.25\%$&Donchess Inference
 & $69.84\%$\\
     \hline
    \end{tabular}
    \caption{10 popular NCAA men's basketball public team ranking systems with their accuracies for 2024 year}
    \label{tab:NCAA}
\end{table}

Lastly, we use the \textquotedblleft ABE" ensemble with score combination to derive rankings for teams. We obtain an average of team 1 winning probabilities for all the games involving that team. The team rankings are obtained by ranking the average probability for all the teams.  Based on this team ranking, we calculate the accuracy, which reaches $71.43\%$. Although it is not the highest accuracy, it beats half of the ranking systems in Table \ref{tab:NCAA}. 

\section{Conclusion and discussion}
\subsection{Conclusion}

Sports prediction is often treated as a classification or regression problem. However, this paper adopts a different perspective by predicting sports outcomes through both rank and score combinations of 5 diverse and fairly good base models. 
The CFA framework generates 56 ensembles from 5 base models.  By examining game results from the previous 10 years, we identified the ensemble model most likely to improve for 2024. We compare our model's accuracy rates against 10 common NCAA basketball prediction websites (Table \ref{tab:NCAA}). The team ranking obtained from the rank combination shows an improvement of $1.58\%$ over the best of the 10 common predictions. In the future, we plan to extend to three types of weights for model combination to explore further improvements in ensemble performance and to compare the results with other ensemble models described in the literature.

\subsection{Discussion}
In this study, we pretrain five relatively diverse five models over the past 10 years to identify the combination that consistently demonstrates improvement. As depicted in Figure \ref{fig:dia}, the CFA framework comprises both score combination and rank combination. Score combination uses scores as inputs, which can be probabilities in classification tasks, or real values in regression tasks. Rank combination, on the other hand, involves their ranks in descending order. 


During model training for each previous year, the output of the score combination is converted into ranks, and then to calculate accuracy. One advantage of the CFA technique is its ability to handle both real values (scores) and discrete values (ranks), making the CFA framework robust and applicable to a wide range of applications.

Another strength of the CFA framework lies in its choice of weights. Three types of weights are utilized: average weight, model performance, and diversity strength derived from cognitive diversity. 
Due to page constraints and space limitations, this study covers only weighted combination by diversity strength as weight using cognitive diversity (Formula \eqref{cd} and \eqref{ds}). Cognitive diversity is a novel diversity measure distinct from traditional correlation methods such as Pearson's correlation, Kendall's tau, and Spearman's rho \cite{hao2018predication}. As shown in Formula (\ref{cd}), cognitive diversity is independent of data items and has been shown to be useful in various domains \cite{hsu2019cognitive}. It can be applied to both supervised and unsupervised learning tasks. In our future work, we will employ all three types of weights, as they each capture the contributions of base models to the resulting ensemble from different perspectives.

Finally, by pretraining a set of five diverse models and using both score and rank combinations, our predictions not only improved the accuracy but also enhanced efficiency.



\footnotesize
\bibliographystyle{ieeetr}
\bibliography{references}

\end{document}